\documentclass{article} %
\usepackage{paper,times}

\usepackage{amsmath,amsfonts,bm}

\def\eqref#1{equation~\ref{#1}}

\def\1{\bm{1}}

\DeclareMathAlphabet{\mathsfit}{\encodingdefault}{\sfdefault}{m}{sl}
\SetMathAlphabet{\mathsfit}{bold}{\encodingdefault}{\sfdefault}{bx}{n}

\newcommand{\E}{\mathbb{E}}

\usepackage{hyperref}
\usepackage{url}
\usepackage{movie15}
\usepackage{animate}
\usepackage{booktabs}
\usepackage{graphicx} %
\usepackage{mathtools}
\usepackage{color, colortbl}

\definecolor{Gray}{gray}{0.8}
\newcommand{\Paragraph}[1]{\vspace{1mm}\noindent\textbf{#1}}

\title{Make-A-Video: Text-to-Video Generation without Text-Video Data}

\author{%
  Uriel Singer  \textsuperscript{+}\\
  \And
  Adam Polyak   \textsuperscript{+}\\
  \And
  Thomas Hayes  \textsuperscript{+}\\
  \And Xi Yin   \textsuperscript{+}\\
  \AND Jie An \\
  \And Songyang Zhang \\
  \And Qiyuan Hu \\
  \And  Harry Yang \\
  \And  Oron Ashual \\
  \And Oran Gafni \AND
  \And Devi Parikh   \textsuperscript{+} \And 
  Sonal Gupta   \textsuperscript{+} \And 
  Yaniv Taigman \textsuperscript{+} \And \\
  \AND \And {Meta AI}
}

\makeatletter
\let\oldfootnote\footnote
\def\footnote{\@ifstar\footnote@star\footnote@nostar}
\def\footnote@star#1{{\let\thefootnote\relax\footnotetext{#1}}}
\def\footnote@nostar{\oldfootnote}
\makeatother

\iclrfinalcopy %
\begin{document}

    \footnote*{\textsuperscript{+} Core Contributors. Corresponding author: \texttt{urielsinger@meta.com}. Jie and Songyang are from University of Rochester (work done during internship at Meta).}

\maketitle

\begin{abstract}

We propose Make-A-Video -- an approach for directly translating the tremendous recent progress in Text-to-Image (T2I) generation to Text-to-Video (T2V). Our intuition is simple: learn what the world looks like and how it is described from paired text-image data, and learn how the world moves from unsupervised video footage. Make-A-Video has three advantages: (1) it accelerates training of the T2V model (it does not need to learn visual and multimodal representations from scratch), (2) it does not require paired text-video data, and (3) the generated videos inherit the vastness (diversity in aesthetic, fantastical depictions, etc.) of today's image generation models. 
We design a simple yet effective way to build on T2I models with novel and effective spatial-temporal modules. First, we decompose the full temporal U-Net and attention tensors and approximate them in space and time. Second, we design a spatial temporal pipeline to generate high resolution and frame rate videos with a video decoder, interpolation model and two super resolution models that can enable various applications besides T2V.
In all aspects, spatial and temporal resolution, faithfulness to text, and quality, Make-A-Video sets the new state-of-the-art in text-to-video generation, as determined by both qualitative and quantitative measures. \\

\end{abstract}

\vspace{-15pt}

\section{Introduction}

\begin{figure}[t]
    \centering
    \includegraphics[trim={60 1525 70 40}, clip, width=\textwidth]{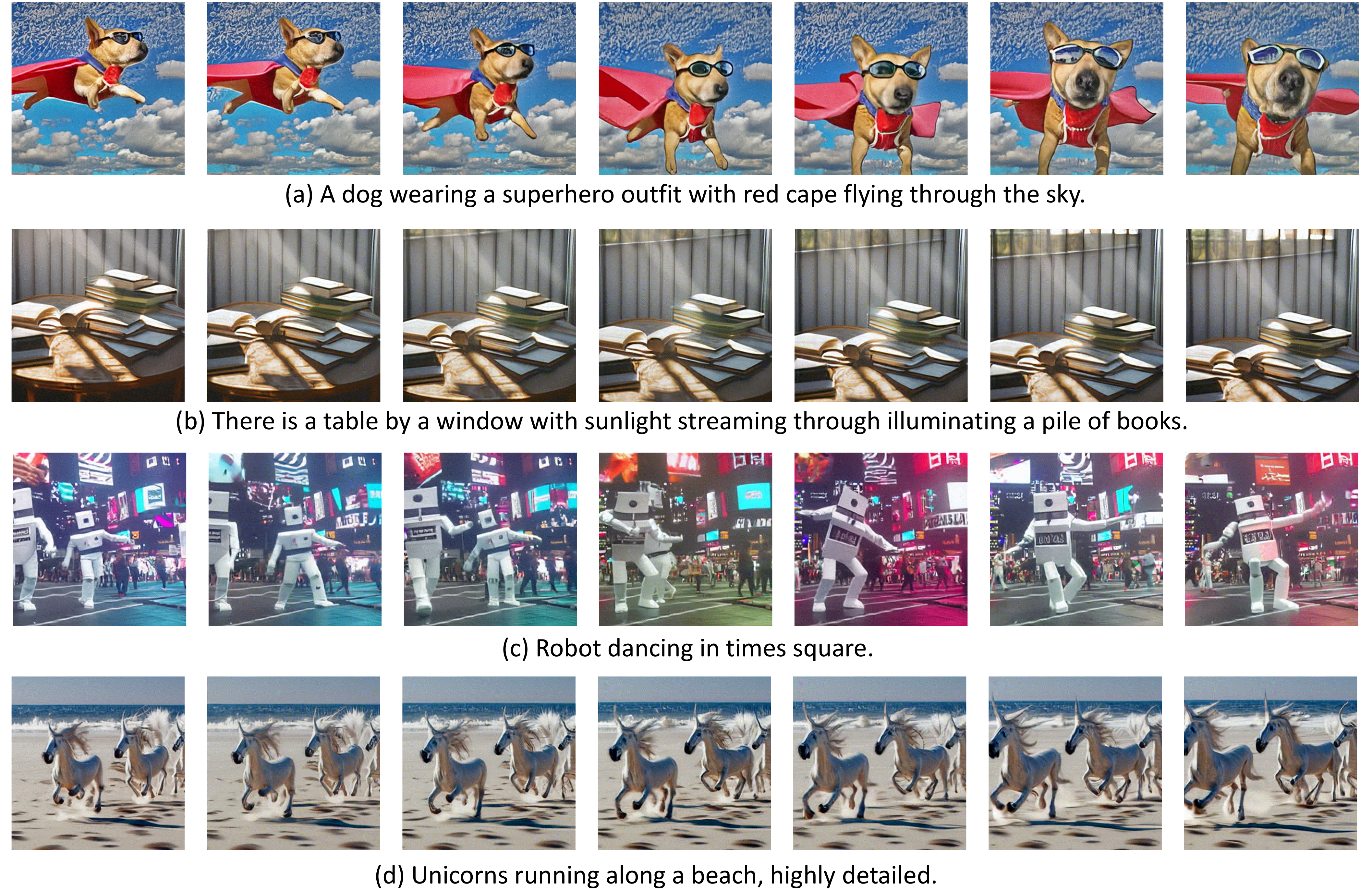}
    \caption{{\textbf{T2V generation examples.}} Our model can generate high-quality videos with coherent motion for a diverse set of visual concepts. In example (a), there are large and realistic motion for the dog. In example (b), the books are almost static but the scene changes with the camera motion. \textbf{Video samples are available at
\href{https://make-a-video.github.io/}{make-a-video.github.io}}
    } 
    \label{fig:illustration}
    \vspace{-10pt}
\end{figure}

The Internet has fueled collecting billions of (alt-text, image) pairs from HTML pages~\citep{laion5b}, enabling the recent breakthroughs in Text-to-Image (T2I) modeling. However, replicating this success for videos is limited since a similarly sized (text, video) dataset cannot be easily collected.
It would be wasteful to train Text-to-Video (T2V) models from scratch
when there already exist models that can generate images.
Moreover, unsupervised learning enables networks to learn from orders of magnitude more data. This large quantity of data is important to learn representations of more subtle, less common concepts in the world. Unsupervised learning has long had great success in advancing the field of natural language processing (NLP)~\citep{roberta, gpt3}. Models pre-trained this way yield considerably higher performance than when solely trained in a supervised manner.

Inspired by these motivations, we propose Make-A-Video. Make-A-Video leverages T2I models to learn the correspondence between text and the visual world, and uses unsupervised learning on unlabeled (unpaired) video data, to learn realistic motion. Together, Make-A-Video generates videos from text without leveraging paired text-video data. 

Clearly, text describing images does not capture the entirety of phenomena observed in videos. That said, one can often infer actions and events from static images (e.g.\ a woman drinking coffee, or an elephant kicking a football)  as done in image-based action recognition systems~\citep{actionsurvey2}. Moreover, even without text descriptions, unsupervised videos are sufficient to learn how different entities in the world move and interact (e.g.\ the motion of waves at the beach, or of an elephant's trunk). As a result, a model that has only seen text describing images is surprisingly effective at generating short videos, as demonstrated by our temporal diffusion-based method. Make-A-Video sets the new state-of-the-art in T2V generation.

Using function-preserving transformations, we extend the spatial layers at the model initialization stage, to include temporal information.  
The extended spatial-temporal network includes new attention modules that learn temporal world dynamics from a collection of videos. This procedure significantly accelerates the T2V training process by instantaneously transferring the knowledge from a previously trained T2I network to a new T2V one. To enhance the visual quality, we train spatial super-resolution models as well as frame interpolation models. This increases the resolution of the generated videos, as well as enables a higher (controllable) frame rate.

Our main contributions are:
\vspace{-5pt}
\begin{itemize}
    \item We present Make-A-Video -- an effective method that extends a diffusion-based T2I model to T2V through a spatiotemporally factorized diffusion model.
    \item We leverage joint text-image priors to bypass the need for paired text-video data, which in turn allows us to potentially scale to larger quantities of video data.
    \item We present super-resolution strategies in space and time that, for the first time, generate high-definition,  high frame-rate videos given a user-provided textual input.
    \item We evaluate Make-A-Video against existing T2V systems and present: (a) State-of-the-art results in quantitative as well as qualitative measures, and (b) A more thorough evaluation than existing literature in T2V. We also collect a test set of 300 prompts for zero-shot T2V human evaluation which we plan to release. %
\end{itemize}

\section{Previous work}
\vspace{-10pt}

\Paragraph{Text-to-Image Generation.}
~\citep{reed2016generative} is among the first methods to extend unconditional Generative Adversairal Network (GAN)~\citep{goodfellow2014generative} to T2I generation. Later GAN variants have focused on progressive generation~\citep{zhang2017stackgan,hong2018inferring}, or better text-image alignment~\citep{xu2018attngan,zhang2021cross}. The pioneering work of DALL-E~\citep{ramesh2021zero} considers T2I generation as a sequence-to-sequence translation problem using a discrete variational auto-encoder (VQVAE) and Transformer~\citep{transformer}. Additional variants~\citep{ding2022cogview2} have been proposed since then. For example, Make-A-Scene~\citep{makeascene} explores controllable T2I generation using semantic maps. Parti~\citep{parti} aims for more diverse content generation through an encoder-decoder architecture and an improved image tokenizer~\citep{yu2021vector}. On the other hand, Denoising Diffusion Probabilistic Models (DDPMs)~\citep{ddpm} are successfully leveraged for T2I generation. GLIDE~\citep{nichol2021glide} trained a T2I and an upsampling diffusion model for cascade generation. GLIDE's proposed classifier-free guidance has been widely adopted in T2I generation to improve image quality and text faithfulness. DALLE-2~\citep{dalle2} leverages the CLIP~\citep{radford2021learning} latent space and a prior model. VQ-diffusion~\citep{gu2022vector} and stable diffusion~\citep{rombach2022high} performs T2I generation in the latent space instead of pixel space to improve efficiency. 

\vspace{-1mm}
\Paragraph{Text-to-Video Generation.}  
While there is remarkable progress in T2I generation, the progress of T2V generation lags behind largely due to two main reasons: the lack of large-scale datasets with high-quality text-video pairs, and the complexity of modeling higher-dimensional video data. Early works~\citep{mittal2017sync,pan2017create,marwah2017attentive,li2018video,gupta2018imagine,liu2019cross} are mainly focused on video generation in simple domains, such as moving digits or specific human actions. To our knowledge, Sync-DRAW~\citep{mittal2017sync} is the first T2V generation approach that leverages a VAE with recurrent attention. \citep{pan2017create} and \citep{li2018video} extend GANs from image generation to T2V generation. 

More recently, GODIVA~\citep{wu2021godiva} is the first to use 2D VQVAE and sparse attention for T2V generation supporting more realistic scenes. NÜWA~\citep{nuwa} extends GODIVA, and presents a unified representation for various generation tasks in a multitask learning scheme. To further improve the performance of T2V generation, CogVideo~\citep{CogVideo} is built on top of a frozen CogView-2~\citep{ding2022cogview2} T2I model by adding additional temporal attention modules. Video Diffusion Models (VDM)~\citep{VDM} uses a space-time factorized U-Net with joint image and video data training. While both CogVideo and VDM collected 10M private text-video pairs for training, our work uses solely open-source datasets, making it easier to reproduce. %

\vspace{-1mm}
\Paragraph{Leveraging Image Priors for Video Generation.}
Due to the complexity of modeling videos and the challenges in high-quality video data collection, it is natural to consider leveraging image priors for videos to simplifying the learning process. After all, an image is a video with a single frame~\citep{bain2021frozen}. In unconditional video generation, MoCoGAN-HD~\citep{tian2021good} formulates video generation as the task of finding a trajectory in the latent space of a pre-trained and fixed image generation model. In T2V generation,  NÜWA~\citep{nuwa} combines image and video datasets in a multitask pre-training stage to improve model generalization for fine-tuning. CogVideo~\citep{CogVideo} uses a pre-trained and fixed T2I model for T2V generation with only a small number of trainable parameters to reduce memory usage during training. But the fixed autoencoder and T2I models can be restrictive for T2V generation. The architecture of VDM~\citep{VDM} can enable joint image and video generation. However, they sample random independent images from random videos as their source of images, and do not leverage the massive text-image datasets. 

Make-A-Video differs from previous works in several aspects. First, our architecture breaks the dependency on text-video pairs for T2V generation.
This is a significant advantage compared to prior work, that has to be restricted to narrow domains~\citep{mittal2017sync,gupta2018imagine,ge2022long,hayes2022mugen}, or require large-scale paired text-video data~\citep{CogVideo,VDM}. 
Second, we fine-tune the T2I model for video generation, gaining the advantage of adapting the model weights effectively, compared to freezing the weights as in CogVideo~\citep{CogVideo}. 
Third, motivated from prior work on efficient architectures for video and 3D vision tasks~\citep{ye20193d,qiu2017learning,xie2018rethinking}, our use of pseudo-3D convolution~\citep{qiu2017learning} and temporal attention layers not only better leverage a T2I architecture, it also allows for better temporal information fusion compared to VDM~\citep{VDM}.

\section{Method}
Make-A-Video consists of three main components: \textbf{(i)} A base T2I model trained on text-image pairs (Sec.~\ref{t2e_model}), \textbf{(ii)} spatiotemporal convolution and attention layers that extend the networks' building blocks to the temporal dimension (Sec.~\ref{spatial_temporal_layers}), and \textbf{(iii)} spatiotemporal networks that consist of both spatiotemporal layers, as well as another crucial element needed for T2V generation - a frame interpolation network for high frame rate generation (Sec.~\ref{frame_interp_network}). 

\begin{figure}[t]

\centering
\includegraphics[width=1\linewidth]{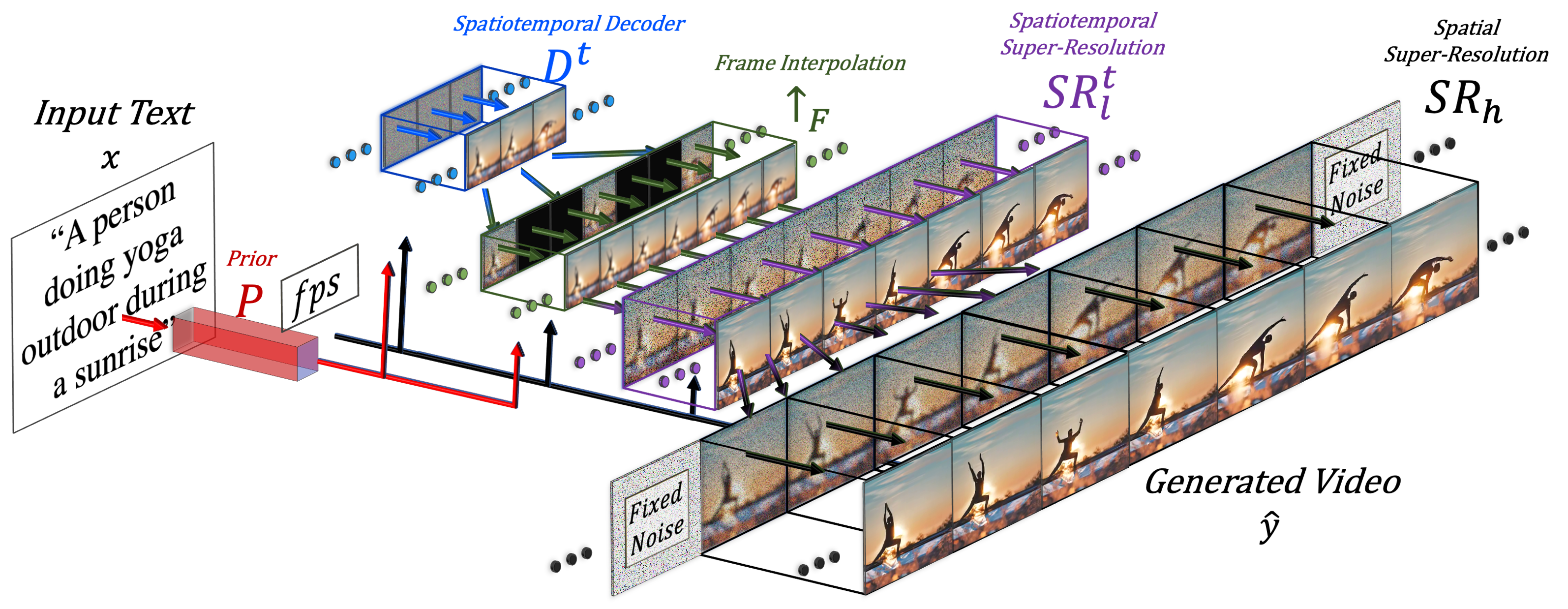} 

\caption{\textbf{Make-A-Video high-level architecture.} Given input text $x$ translated by the prior $\operatorname{P}$ into an image embedding, and a desired frame rate $fps$, the decoder $\operatorname{D^t}$ generates $16$ $64\times64$ frames, which are then interpolated to a higher frame rate by $\uparrow_{F}$, and increased in resolution to $256\times256$ by $\operatorname{SR}_l^t$ and $768\times768$ by $\operatorname{SR}_h$, resulting in a high-spatiotemporal-resolution generated video $\hat{y}$.
} 
\label{fig:arch_highlevel}
\vspace{-10pt}
\end{figure}

Make-A-Video's final T2V inference scheme (depicted in Fig.~\ref{fig:arch_highlevel}) can be formulated as:
\begin{equation}
\hat{y_t}=\operatorname{SR}_h\circ\operatorname{SR}_l^t\circ \uparrow_{F}\circ\operatorname{D}^t\circ\operatorname{P}\circ(\hat{x},\operatorname{C}_x(x)),
\label{eq_inference}
\end{equation}
where $\hat{y_t}$ is the generated video, $\operatorname{SR}_h,\operatorname{SR}_l$ are the spatial and spatiotemporal super-resolution networks (Sec.~\ref{spatial_temporal_layers}), $\uparrow_{F}$ is a frame interpolation network (Sec.~\ref{frame_interp_network}), $\operatorname{D}^t$ is the spatiotemporal decoder (Sec.~\ref{spatial_temporal_layers}), $\operatorname{P}$ is the prior (Sec.~\ref{t2e_model}), $\hat{x}$ is the BPE-encoded text, $\operatorname{C}_x$ is the CLIP text encoder~\citep{radford2021learning}, and $x$ is the input text. The three main components are described in detail in the following sections.

\newcommand{\bx}{\mathbf{x}}
\newcommand{\Eb}[2]{\E_{#1}\!\left[#2\right]}
\newcommand{\bepsilon}{{\boldsymbol{\epsilon}}}
\newcommand{\bc}{\mathbf{c}}
\newcommand{\bI}{\mathbf{I}}
\newcommand{\bzero}{\mathbf{0}}
\newcommand{\bz}{\mathbf{z}}

\vspace{-5pt}
\subsection{Text-to-image model}
\vspace{-5pt}
\label{t2e_model}
Prior to the addition of the temporal components, we train the backbone of our method: a T2I model trained on text-image pairs, sharing the core components with the work of~\citep{dalle2}. 

We use the following networks to produce high-resolution images from text:
\textbf{(i) A prior network} $\operatorname{\textbf{P}}$, that during inference generates image embeddings $y_e$ given text embeddings $x_e$ and BPE encoded text tokens $\hat{x}$, \textbf{(ii) a decoder network} $\operatorname{\textbf{D}}$ that generates a low-resolution $64\times64$ RGB image $\hat{y}_l$, conditioned on the image embeddings $y_e$, and \textbf{(iii) two super-resolution networks} $\operatorname{\textbf{SR}}_\textbf{l}$,$\operatorname{\textbf{SR}}_\textbf{h}$ that increase the generated image $\hat{y}_l$ resolution to $256\times256$ and $768\times768$ pixels respectively, resulting in the final\footnote{We then downsample to 512 using bicubic interpolation for a cleaner aesthetic. Maintaining a clean aesthetic for high definition videos is part of future work.} generated image $\hat{y}$.

\begin{figure}[t]
\begin{tabular}{cc}
\centering
\includegraphics[height=4.2cm]{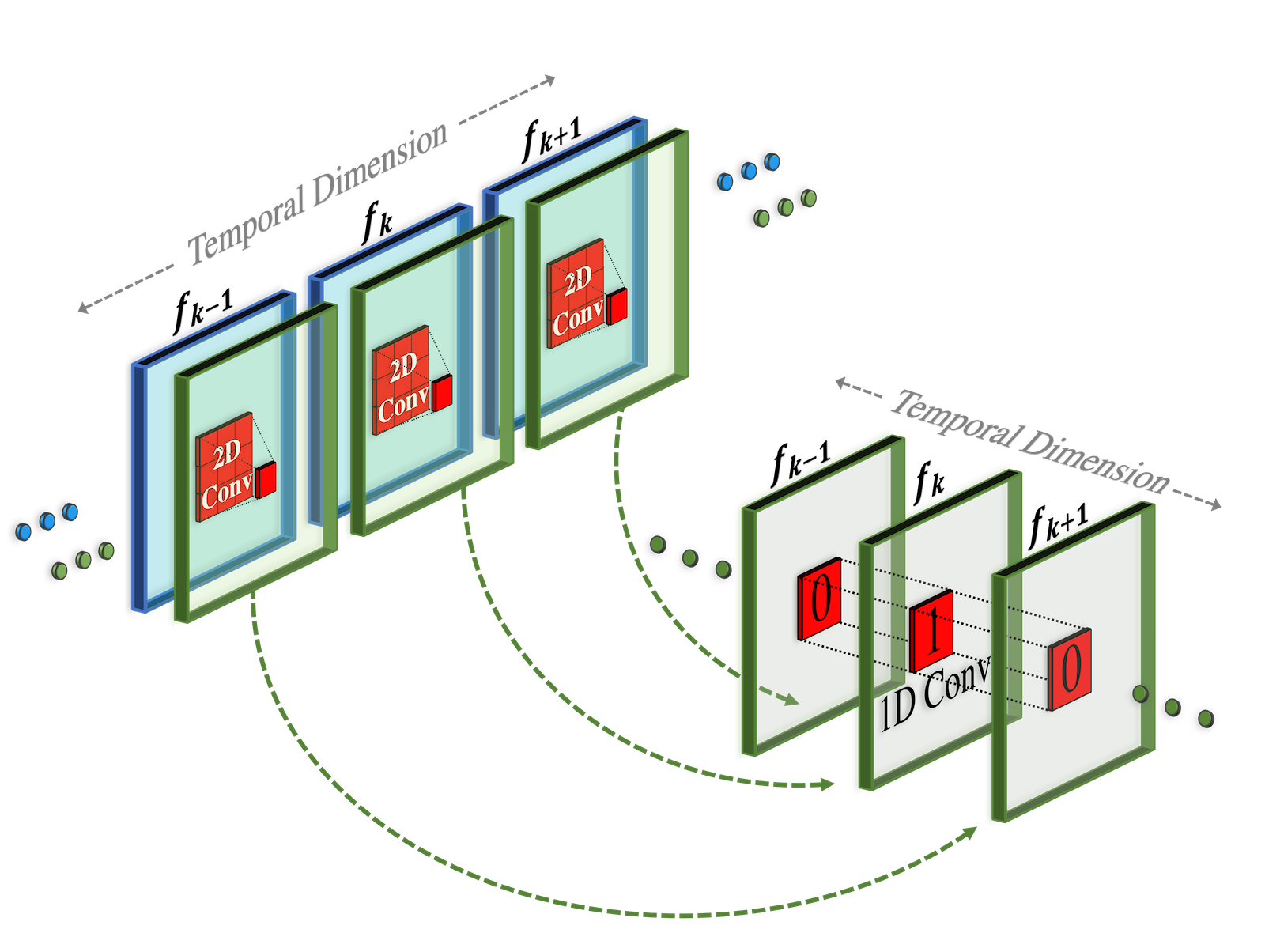} & \includegraphics[height=4.7cm]{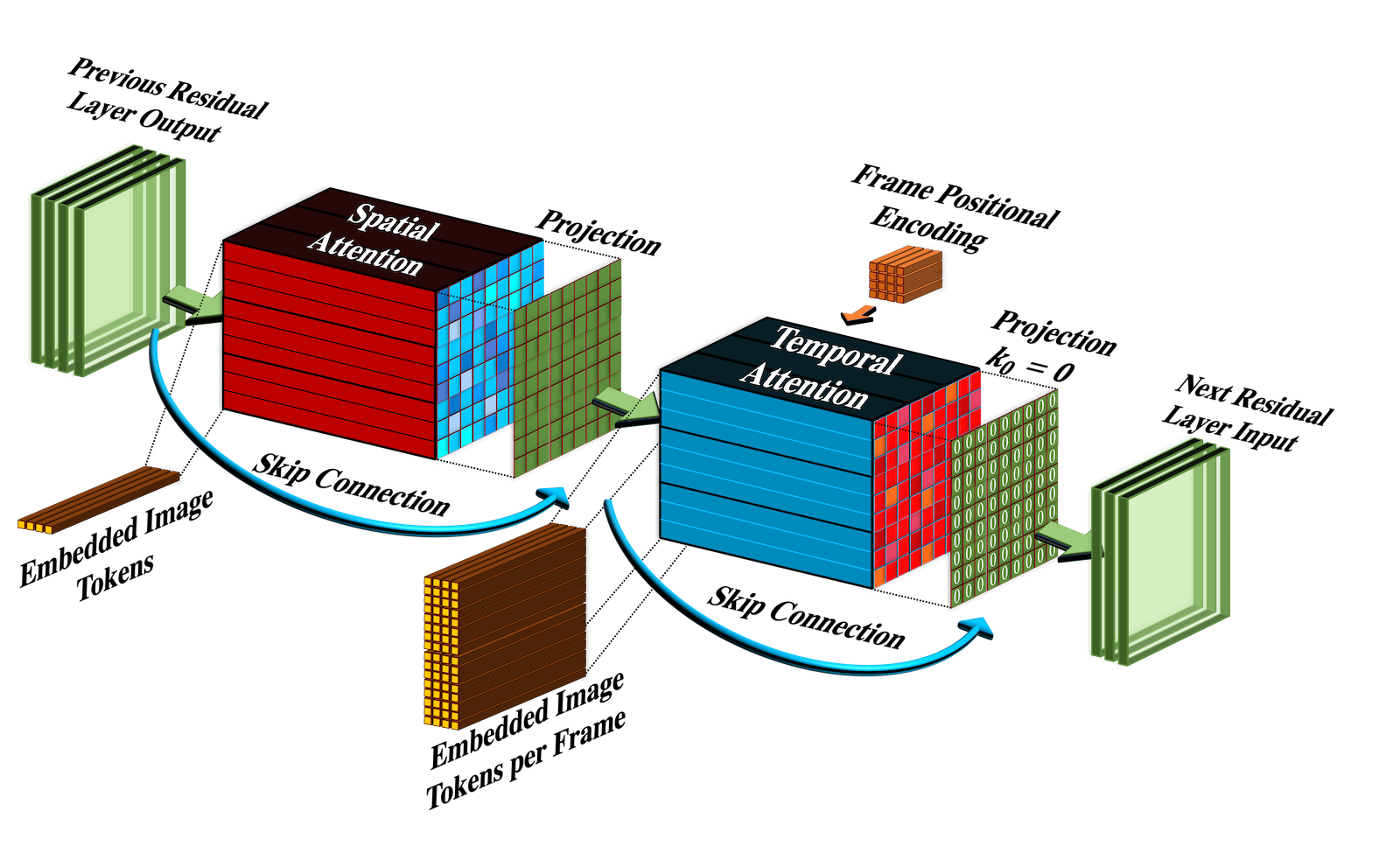} %
\end{tabular}
\vspace{-10pt}
\caption{\textbf{The architecture and initialization scheme of the Pseudo-3D convolutional and attention layers, enabling the seamless transition of a pre-trained Text-to-Image model to the temporal dimension. (left)} Each spatial 2D conv layer is followed by a temporal 1D conv layer. The temporal conv layer is initialized with an identity function. \textbf{(right)} Temporal attention layers are applied following the spatial attention layers by initializing the temporal projection to zero, resulting in an identity function of the temporal attention blocks.}
\label{fig:arch_conv2p5}
\vspace{-10pt}
\end{figure}

\vspace{-5pt}
\subsection{Spatiotemporal layers}
\vspace{-5pt}
\label{spatial_temporal_layers}
In order to expand the two-dimensional (2D) conditional network into the temporal dimension, we modify the two key building blocks that now require not just spatial but also temporal dimensions in order to generate videos: (i) Convolutional layers (Sec.~\ref{2p5d_conv}), and (ii) attention layers (Sec.~\ref{2p5d_atten}), discussed in the following two subsections. Other layers, such as fully-connected layers, do not require specific handling when adding an additional dimension, as they are agnostic to structured spatial and temporal information.
Temporal modifications are made in most U-Net-based diffusion networks: the spatiotemporal decoder $\operatorname{D^t}$ now generating $16$ RGB frames, each of size $64\times64$, the newly added frame interpolation network $\uparrow_{F}$, increasing the effective frame rate by interpolating between the $16$ generated frames (as depicted in Fig.~\ref{fig:arch_highlevel}), and the super-resolution networks $\operatorname{SR}_l^t$. %

Note that super resolution involves hallucinating information. In order to not have flickering artifacts, the hallucination must be consistent across frames. As a result, our $\operatorname{SR}_l^t$ module operates across spatial and temporal dimensions. In qualitative inspection we found this to significantly outperform per-frame super resolution. It is challenging to extend $\operatorname{SR}_h$ to the temporal dimension due to memory and compute constraints, as well as a scarcity of high resolution video data. So $\operatorname{SR}_h$ operates only along the spatial dimensions. But to encourage consistent detail hallucination across frames, we use the same noise initialization for each frame.

\subsubsection{Pseudo-3D convolutional layers}
\label{2p5d_conv}
Motivated by separable convolutions~\citep{chollet2017xception}, we stack a 1D convolution following each 2D convolutional (conv) layer, as shown in Fig.~\ref{fig:arch_conv2p5}. This facilitates information sharing between the spatial and temporal axes, without succumbing to the heavy computational load of 3D conv layers. In addition, it creates a concrete partition between the pre-trained 2D conv layers and the newly initialized 1D conv layers, allowing us to train the temporal convolutions from scratch, while retaining the previously learned spatial knowledge in the spatial convolutions' weights.

Given an input tensor $h \in \mathbb{R}^{B \times C \times F \times H \times W}$, where $B$, $C$, $F$, $H$, $W$ are the batch, channels, frames, height, and width dimensions respectively, the Pseudo-3D convolutional layer is defined as:

\vspace{-10pt}
\begin{equation}
Conv_{P3D} (h) := Conv_{1D}(Conv_{2D}(h) \circ T) \circ T,
\end{equation}
\vspace{-10pt}

where the transpose operator $\circ T$ swaps between the spatial and temporal dimensions.
For smooth initialization, while the $Conv_{2D}$ layer is initialized from the pre-trained T2I model, the $Conv_{1D}$ layer is initialized as the identity function, enabling a seamless transition from training spatial-only layers, to spatiotemporal layers. Note that at initialization, the network will generate K different images (due to random noise), each faithful to the input text but lacking temporal coherence.

\subsubsection{Pseudo-3D attention layers}
\label{2p5d_atten}
A crucial component of T2I networks is the attention layer, where in addition to self-attending to extracted features, text information is injected to several network hierarchies, alongside other relevant information, such as the diffusion time-step. While using 3D convolutional layers is computationally heavy, adding the temporal dimension to attention layers is outright infeasible in terms of memory consumption.
Inspired by the work of~\citep{VDM}, we extend our dimension decomposition strategy to attention layers as well. Following each (pre-trained) spatial attention layer, we stack a temporal attention layer, which   
as with the convolutional layers, approximates a full spatiotemporal attention layer. Specifically, given an input tensor $h$, we define $flatten$ as a matrix operator that flattens the spatial dimension into $h' \in R^{B \times C \times F \times HW}$. $unflatten$ is defined as the inverse matrix operator. The Pseudo-3D attention layer therefore is therefore defined as:

\begin{equation}
ATTN_{P3D}(h) = unflatten(ATTN_{1D}(ATTN_{2D}(flatten(h)) \circ T) \circ T).
\end{equation}

Similarly to $Conv_{P3D}$, to allow for smooth spatiotemporal initialization, the $ATTN_{2D}$ layer is initialized from the pre-trained T2I model and the $ATTN_{1D}$ layer is initialized as the identity function.

Factorized space-time attention layers have also been used in VDM~\citep{VDM} and CogVideo~\citep{CogVideo}. CogVideo has added temporal layers to each (frozen) spatial layers whereas we train them jointly. In order to force their network to train for images and videos interchangeably, VDM has extended their 2D U-Net to 3D through unflattened 1x3x3 convolution filters, such that the subsequent spatial attention remains 2D, and added 1D temporal attention through relative position embeddings. In contrast, we apply an additional 3x1x1 convolution projection (after each 1x3x3) such that the temporal information will also be passed through each convolution layer.%

\noindent \textbf{Frame rate conditioning.} In addition to the T2I conditionings, similar to CogVideo~\citep{CogVideo}, we add an additional conditioning parameter $fps$, representing the number of frames-per-second in a generated video. Conditioning on a varying number of frames-per-second, enables an additional augmentation method to tackle the limited volume of available videos at training time, and provides additional control on the generated video at inference time.

\vspace{-10pt}
\subsection{Frame interpolation network} 
\vspace{-5pt}
\label{frame_interp_network}
In addition to the spatiotemporal modifications discussed in Sec.~\ref{spatial_temporal_layers}, we train a new masked frame interpolation and extrapolation network $\uparrow_{F}$, capable of increasing the number of frames of the generated video either by frame interpolation for a smoother generated video, or by pre/post frame extrapolation for extending the video length.
In order to increase the frame rate within memory and compute constraints, we 
fine-tune a spatiotemporal decoder $\operatorname{D^t}$ on the task of masked frame interpolation, by zero-padding the masked input frames, enabling video upsampling. When fine-tuning on masked frame interpolation, we add an additional 4 channels to the input of the U-Net: 3 channels for the RGB masked video input and an additional binary channel indicating which frames are masked. We fine-tune with variable frame-skips and $fps$ conditioning to enable multiple temporal upsample rates at inference time. We denote $\uparrow_{F}$ as the operator that expands the given video tensor through masked frame interpolation. For all of our experiments we applied $\uparrow_{F}$ with frame skip 5 to upsample a 16 frame video to 76 frames ((16-1)$\times$5+1). Note that we can use the same architecture for video extrapolation or image animation by masking frames at the beginning or end of a video.

\vspace{-10pt}
\subsection{Training} %
The different components of Make-A-Video described above are trained independently. The only component that receives text as input is the prior $\operatorname{P}$. We train it on paired text-image data and do not fine-tune it on videos. The decoder, prior, and two super-resolution components are first trained on images alone (no aligned text). Recall that the decoder receives CLIP image embedding as input, and the super-resolution components receive downsampled images as input during training. 
After training on images, we add and initialize the new temporal layers and fine-tune them over {unlabeled} video data.
16 frames are sampled from the original video with random $fps$ ranging from $1$ to $30$. We use the beta function for sampling and while training the decoder, start from higher FPS ranges (less motion) and then transition to lower FPS ranges (more motion). 
The masked-frame-interpolation component is fine-tuned from the temporal decoder.

\section{Experiments}

\vspace{-5pt}
\subsection{Datasets and Settings}
\vspace{-5pt}

\Paragraph{Datasets.} 
To train the image models, we use a $2.3$B subset of the dataset from
~\citep{schuhmannlaion} where the text is English. We filter out sample pairs with NSFW images~\footnote{We used this model: https://github.com/GantMan/nsfw\_model}, toxic words in the text, or images with a watermark probability larger than $0.5$. 
We use  WebVid-10M~\citep{bain2021frozen} and a $10$M subset from HD-VILA-100M~\citep{xue2022advancing}~\footnote{These $100$M clips are sourced from $3.1$M videos. We randomly downloaded $3$ clips per video to form our HD-VILA-10M subset.} to train our video generation models.
Note that only the videos (no aligned text) are used. The decoder $\operatorname{D}^t$ and the interpolation model is trained on WebVid-10M. $\operatorname{SR}_l^t$ is trained on both WebVid-10M and HD-VILA-10M. 
While prior work~\citep{CogVideo,VDM} have collected private text-video pairs for T2V generation, we use only public datasets (and no paired text for videos). We conduct automatic evaluation on UCF-101~\citep{soomro2012ucf101} and MSR-VTT~\citep{xu2016msr} in a zero-shot setting.

\begin{table}[t]
\caption{T2V generation evaluation on MSR-VTT. Zero-Shot means no training is conducted on MSR-VTT. Samples/Input means how many samples are generated (and then ranked) for each input.}
\label{tab:eval_msrvtt}
\centering
\begin{tabular}{@{}lcc|cc@{}}
\toprule
Method & Zero-Shot & Samples/Input & FID ($\downarrow$) & CLIPSIM ($\uparrow$) \\
GODIVA~\citep{wu2021godiva} & No & $30$ & $-$ & $0.2402$ \\
NÜWA~\citep{nuwa} & No & $-$ & $47.68$ & $0.2439$ \\
CogVideo~\citep{CogVideo} (Chinese) & Yes & $1$ & $24.78$ & $0.2614$ \\
CogVideo~\citep{CogVideo} (English) & Yes & $1$ & $23.59 $ & $0.2631$ \\
\toprule
\rowcolor{Gray}
Make-A-Video (ours) & Yes & $1$ & $\textbf{13.17}$ & $\textbf{0.3049}$ \\
\bottomrule
\end{tabular}
\end{table}

\Paragraph{Automatic Metrics.}
For UCF-101, we write one template sentence for each class (without generating any video) and fix it for evaluation. We report Frechet Video Distance (FVD) and Inception Score (IS) on $10$K samples following~\citep{VDM}. We generate samples that follow the same class distribution as the training set. For MSR-VTT, we report Frechet Inception Distance (FID)~\citep{parmar2021cleanfid} and CLIPSIM (average CLIP similarity between video frames and text)~\citep{wu2021godiva}, where all $59,794$ captions from the test set are used, following~\citep{nuwa}.

\Paragraph{Human Evaluation Set and Metrics.}
We collect an evaluation set from Amazon Mechanical Turk (AMT) that consists of $300$ prompts. We asked annotators what they would be interested in generating if there were a T2V system. We filtered out prompts that were incomplete (e.g., ``jump into water"), too abstract (e.g., ``climate change"), or offensive. We then identified $5$ categories (animals, fantasy, people, nature and scenes, food and beverage) and selected prompts for these categories. These prompts were selected without generating any videos for them, and were kept fixed. In addition, we also used the DrawBench prompts from Imagen~\citep{imagen} for human evaluation. 
We evaluate video quality and text-video faithfulness. For video quality, we show two videos in random order and ask annotators which one is of higher quality. For faithfulness, we additionally show the text and ask annotators which video has a better correspondence with the text (we suggest them to ignore quality issues). In addition, we also conducted human evaluation to compare video motion realism of our interpolation model and FILM~\citep{reda2022film}.
For each comparison, we use the majority vote from $5$ different annotators as the final result.

\vspace{-5pt}
\subsection{Quantitative Results}
\vspace{-5pt}

\Paragraph{Automatic Evaluation on MSR-VTT.} In addition to GODIVA and NÜWA that report on MSR-VTT, we also perform inference on the officially released CogVideo model with both Chinese and English inputs for comparison. For CogVideo and Make-A-Video, we only generate one sample for each prompt in a zero-shot setting. We only generate videos that are at $16\times256\times256$ as the evaluation models do not expect higher resolutions and frame rate. 
The results are shown in Table~\ref{tab:eval_msrvtt}. Make-A-Video's zero-shot performance is much better than GODIVA and NÜWA which are trained on MSR-VTT. We also outperform CogVideo in both Chinese and English settings. Thus, Make-A-Video has significantly better generalization capabilities than prior work.

\begin{table}[t]
\caption{Video generation evaluation on UCF-101 for both zero-shot and fine-tuning settings.} %
\label{tab:eval_ucf}
\centering
\begin{tabular}{lccc|cc}
\toprule
Method & Pretrain & Class & Resolution & IS ($\uparrow$) & FVD ($\downarrow$)\\
\midrule
\multicolumn{6}{c}{Zero-Shot Setting} \\
\midrule
CogVideo (Chinese) & No & Yes & $480\times480$ & $23.55$ & $751.34$ \\
CogVideo (English) & No & Yes & $480\times480$ & $25.27$ & $701.59$ \\
\rowcolor{Gray}
Make-A-Video (ours) & No & Yes & $256\times256$ & $\bf 33.00$ & $\bf 367.23$ \\
\midrule
\multicolumn{6}{c}{Finetuning Setting} \\
\midrule
TGANv2\citep{saito2020train} & No & No & $128\times128$ & $26.60\pm0.47$ & - \\
DIGAN\citep{yu2022generating} & No & No & & $32.70\pm0.35$ & $577\pm22$\\
MoCoGAN-HD\citep{tian2021good} & No & No & $256\times256$ & $33.95\pm0.25$ & $700\pm24$  \\
CogVideo~\citep{CogVideo} & Yes & Yes & $160\times160$ & $50.46$ & $626$\\
VDM~\citep{VDM} & No & No & $64\times64$ & $57.80\pm1.3$ & - \\
TATS-base\citep{ge2022long} & No & Yes & $128\times128$ & $79.28\pm0.38$ & $278\pm11$\\
\midrule
\rowcolor{Gray}
Make-A-Video (ours)  & Yes & Yes & $256\times256$ & $\textbf{82.55}$  & $\textbf{81.25}$ \\
\bottomrule
\end{tabular}
\end{table}

\begin{table}[t]
\caption{Human evaluation results compared to CogVideo~\citep{CogVideo} on DrawBench and our test set, and to VDM~\citep{VDM} on the $28$ examples from their website. The numbers show the percentage of raters that prefer the results of our Make-A-Video model.}
\label{tab:eval_human}
\centering
\begin{tabular}{lc|cc}
\toprule
Comparison & Benchmark & Quality & Faithfulness \\
\toprule
Make-A-Video (ours) vs. VDM & VDM prompts ($28$) & $84.38$ & $78.13$ \\
Make-A-Video (ours) vs. CogVideo (Chinese) & DrawBench ($200$) & $76.88$ & $73.37$ \\ 
Make-A-Video (ours) vs. CogVideo (English) & DrawBench ($200$) & $74.48$ & $68.75$\\ 
Make-A-Video (ours) vs. CogVideo (Chinese) & Our Eval. Set ($300$) & $73.44$ & $75.74$ \\ 
Make-A-Video (ours) vs. CogVideo (English) & Our Eval. Set ($300$) & $77.15$ & $71.19$\\ 
\bottomrule
\end{tabular}
\end{table}

\Paragraph{Automatic Evaluation on UCF-101.} UCF-101 is a popular benchmark to evaluate video generation and has been recently used in T2V models. CogVideo performed finetuning of their pretrained model for class-conditional video generation. VDM~\citep{VDM} performed unconditional video generation and trained from scratch on UCF-101. We argue that both settings are not ideal and is not a direct evaluation of the T2V generation capabilities. Moreover, the FVD evaluation model expects the videos to be $0.5$ second ($16$ frames), which is too short to be used for video generation in practice. Nevertheless, in order to compare to prior work, we conducted evaluation on UCF-101 in both zero-shot and finetuning settings. 
As shown in Table~\ref{tab:eval_ucf}, Make-A-Video's zero-shot performance is already competitive than other approaches that are trained on UCF-101, and is much better than CogVideo, which indicates that Make-A-Video can generalize better even to such a specific domain. Our finetuning setting achieves state-of-the-art results with a significant reduction in FVD, which suggests that Make-A-Video can generate more coherent videos than prior work.

\vspace{-5pt}
\begin{figure}[t]
    \centering
    \includegraphics[trim={30 350 20 20},clip,width=1\textwidth]{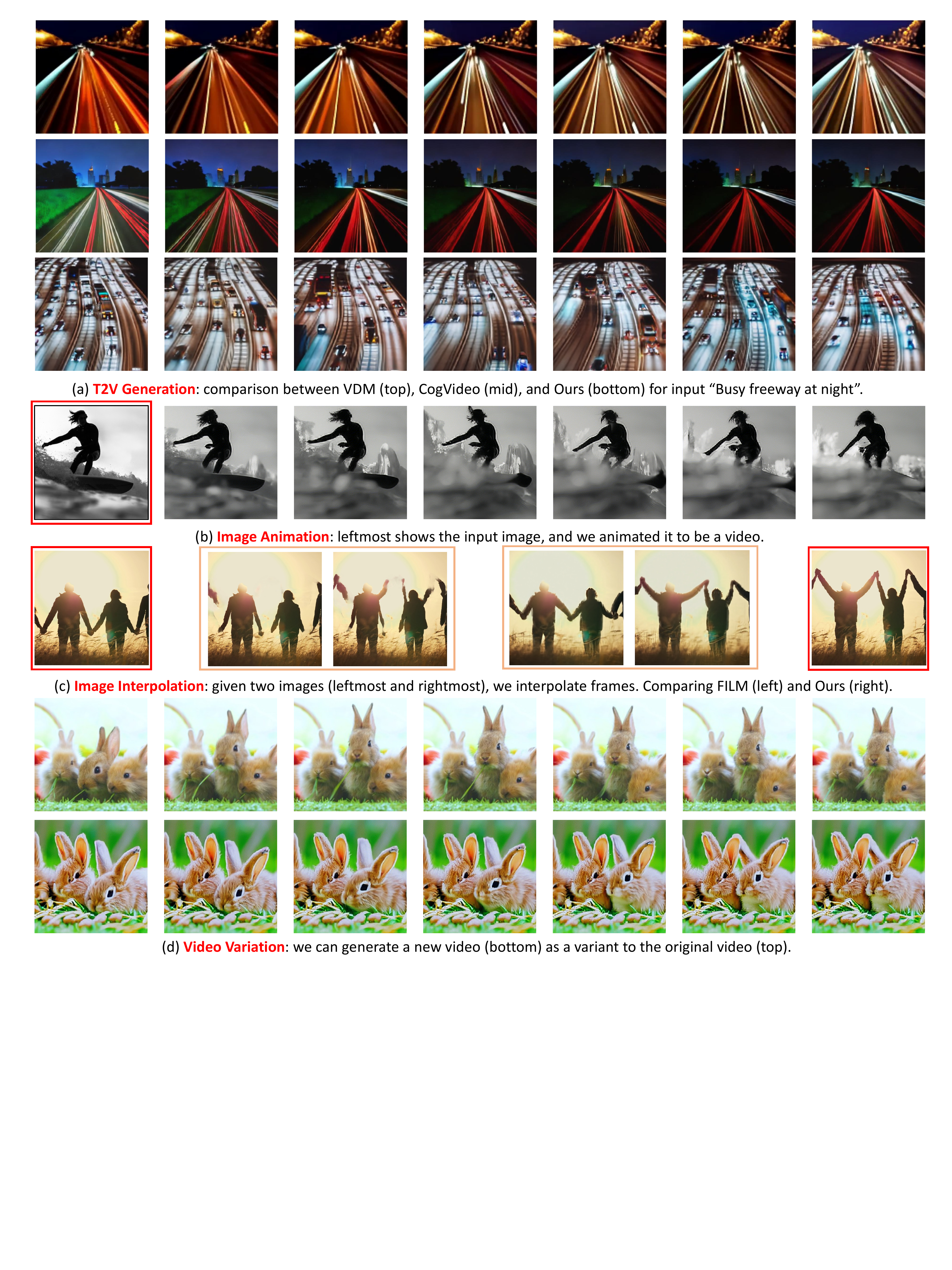}
    \vspace{-10pt}
    \caption{Qualitative results for various comparisons and applications.} %
    \label{fig:visual_results}
    \vspace{-10pt}
\end{figure}

\Paragraph{Human Evaluation.}
We compare to CogVideo (the only public zero-shot T2V generation model) on  DrawBench and our test set. We also evaluate on the $28$ videos shown on the webpage of VDM~\citep{VDM} (which may be biased towards showcasing the model's strengths). Since this is a very small test set, we randomly generate $8$ videos for each input and perform evaluation $8$ times and report the average results. 
We generate videos at $76\times256\times256$ resolution for human evaluation. 
The results are shown in Table~\ref{tab:eval_human}. Make-A-Video achieves much better performance in both video quality and text-video faithfulness in all benchmarks and comparisons. For CogVideo, the results are similar on DrawBench and our evaluation set. For VDM, it is worth noting that we have achieved significantly better results 
without any cherry-picking. 
We also evaluate our frame interpolation network in comparison to FILM~\citep{reda2022film}. We first generate low frame rate videos (1 FPS) from text prompts in DrawBench and our evaluation set, then use each method to upsample to 4 FPS. Raters choose our method for more realistic motion 62\% of the time on our evaluation set and 54\% of the time on DrawBench. We observe that our method excels when there are large differences between frames where having real-world knowledge of how objects move is crucial.

\vspace{-5pt}
\subsection{Qualitative Results}
\vspace{-5pt}
Examples of Make-A-Video's  generations are shown in Figure~\ref{fig:illustration}. In this section, we will show T2V generation comparison to CogVideo~\citep{CogVideo} and VDM~\citep{VDM}, and video interpolation comparison to FILM~\citep{reda2022film}. 
In addition, our models can be used for a variety of other tasks such as image animation, video variation, etc. Due to space constraint, we only show a single example of each. 
Figure~\ref{fig:visual_results} (a) shows the comparison of Make-A-Video to CogVideo and VDM. Make-A-Video can generate richer content with motion consistency and text correspondence.
Figure~\ref{fig:visual_results} (b) shows an example of image animation where we condition the masked frame interpolation and extrapolation network $\uparrow_{F}$ on the image and CLIP image embedding to extrapolate the rest of the video. %
This allows a user to generate a video using their own image -- giving them the opportunity to personalize and directly control the generated video. Figure~\ref{fig:visual_results} (c) shows a comparison of our approach to FILM~\citep{reda2022film} on the task of interpolation between two images. We achieve this by using the interpolation model that takes the two images as the beginning and end frames and masks $14$ frames in between for generation. Our model generates more semantically meaningful interpolation while FILM seems to primarily smoothly transition between frames without semantic real-world understanding of what is moving. Figure~\ref{fig:visual_results} (d) shows an example for video variation. We take the average CLIP embedding of all frames from a video as the condition to generate a semantically similar video. 
More video generation examples and  applications can be found here: \href{https://make-a-video.github.io/}{make-a-video.github.io}.

\if, 0
\paragraph{T2V Generation.} 
Figure~\ref{} shows the qualitative results of Make-A-Video compared to CogVideo. We observe that our model can generate much better results with motion consistency and text correspondence. 

\paragraph{Image Animation.}
In addition to T2V generation, Make-A-Video can also be used to perform image animation. Specifically, we can use the CLIP image encoder to generate the image embedding as the first frame condition and use our decoder to generate a video. Some results are shown in Figure~\ref{}. \devi{This allows a user to generate a short video for their own image -- giving the user the opportunity to personalize and directly control the generated video.} \devi{Refer to webpage for examples.}

\paragraph{Interpolation between two images.}
In similar vein, our model can... \devi{Describe what and how. Refer to webpage for examples}.

\paragraph{Video extrapolation}
\devi{Describe what and how. Talk about how repeated application of this gives longer videos. Refer to webpage for videos that are 24 seconds long.}

\paragraph{Variations of a video}
\devi{Describe what and how.}

\devi{Any other tasks we get ``for free''?}

\paragraph{Intermediate Results? 16x256x256 to 16x720x720 to 64x720x720}
Figure~\ref{} shows the intermediate results of our model. \devi{760 and not 720?}

\devi{Make this point somewhere: FILM seems to just smoothly transition between frames without semantic real-world understanding of what's moving.}

\devi{Make sure we are referring to all Figures and Tables in the text somewhere.}
\fi

\if, 0
\begin{itemize}
    \item  FVD: \url{https://github.com/google-research/google-research/blob/master/frechet_video_distance/frechet_video_distance.py}
    \item Comparisons to CogVideo; Baseline code: \url{https://github.com/THUDM/CogVideo}
    \item Evaluation scripts (FID/FVD) on: UCF-101 \& MSR-VTT
    \item Animated Drawbench; Extract sentences with verbs
    \item Others less relevant from VDM paper: BAIR Robot Pushing, Kinetics-600
\end{itemize}
\fi

\if, 0
\begin{table}
    \caption{Left: UCF-101, Right: Kintecics-600}
  \label{tab:machine eval}
  \centering
  \begin{minipage}{0.47\columnwidth}
  \centering
  \begin{tabular}{lcc}
    \toprule
    Method & IS ($\uparrow$) & FVD ($\downarrow$)\\
    \midrule

    VideoGPT\citep{yan2021videogpt}  &  24.69 & - \\
    DVD-GAN\citep{clark2019adversarial} & 27.38 & - \\
    TGANv2\citep{saito2020train}* & 28.87 & 1209 \\
    MoCoGAN-HD\citep{tian2021good} & 32.36 & 838  \\
    DIGAN\citep{yu2022generating}* & 29.71  & 655 \\
    DIGAN\citep{yu2022generating} & 32.70  & 577\\
    TATS-base\citep{ge2022long} & 79.28  & 332\\
    CogVideo  & 50.46  & 626\\
    \midrule
    Ours  & NA & NA\\
    \bottomrule
  \end{tabular}
\end{minipage}
\begin{minipage}{0.47\columnwidth}
  \centering
  \begin{tabular}{lc}
    \toprule
    Method & FVD($\downarrow$) \\
    \midrule
    Latent Video Tranformer\citep{rakhimov2020latent} & 224.73    \\
    Video Transformer\citep{weissenborn2019scaling}     & 170 \\
    DVD-GAN-FP\citep{clark2019adversarial}     & 69.15  \\%
    TriVD-GAN-FP\citep{luc2020transformation}  & 25.74  \\
    CogVideo & 109.23 \\
    \midrule
    Ours & NA \\
    \bottomrule
  \end{tabular}
  
  \end{minipage}
\end{table}
\fi

\vspace{-10pt}
\section{Discussion}
\vspace{-10pt}
Learning from the world around us is one of the greatest strengths of human intelligence. Just as we quickly learn to recognize people, places, things, and actions through observation, generative systems will be more creative and useful if they can mimic the way humans learn. Learning world dynamics from orders of magnitude more videos using unsupervised learning helps researchers break away from the reliance on labeled data. The presented work has shown how labeled images combined effectively with unlabeled video footage can achieve that. 

As a next step we plan to address several of the technical limitations. As discussed earlier, our approach can not learn associations between text and phenomenon that can only be inferred in videos. How to incorporate these (e.g., generating a video of a person waving their hand left-to-right or right-to-left), along with generating longer videos, with multiple scenes and events, depicting more detailed stories, is left for future work. 

As with all large-scale models trained on data from the web, our models have learnt and likely exaggerated social biases, including harmful ones. Our T2I generation model was trained on data that removed NSFW content and toxic words. All our data (image as well as videos) is publicly available, adding a layer of transparency to our models, and making it possible for the community to reproduce our work.

\label{headings}

\subsubsection*{Acknowledgments}
Mustafa Said Mehmetoglu, Jacob Xu, Katayoun Zand, Jia-Bin-Huang, Jiebo Luo, Shelly Sheynin, Angela Fan, Kelly Freed. Thank you for your contributions!

\bibliography{paper}

\begin{thebibliography}{49}
\providecommand{\natexlab}[1]{#1}
\providecommand{\url}[1]{\texttt{#1}}
\expandafter\ifx\csname urlstyle\endcsname\relax
  \providecommand{\doi}[1]{doi: #1}\else
  \providecommand{\doi}{doi: \begingroup \urlstyle{rm}\Url}\fi

\bibitem[Bain et~al.(2021)Bain, Nagrani, Varol, and Zisserman]{bain2021frozen}
Max Bain, Arsha Nagrani, G{\"u}l Varol, and Andrew Zisserman.
\newblock Frozen in time: A joint video and image encoder for end-to-end
  retrieval.
\newblock In \emph{ICCV}, pp.\  1728--1738, 2021.

\bibitem[Brown et~al.(2020)Brown, Mann, Ryder, Subbiah, Kaplan, Dhariwal,
  Neelakantan, Shyam, Sastry, Askell, Agarwal, Herbert{-}Voss, Krueger,
  Henighan, Child, Ramesh, Ziegler, Wu, Winter, Hesse, Chen, Sigler, Litwin,
  Gray, Chess, Clark, Berner, McCandlish, Radford, Sutskever, and Amodei]{gpt3}
Tom~B. Brown, Benjamin Mann, Nick Ryder, Melanie Subbiah, Jared Kaplan,
  Prafulla Dhariwal, Arvind Neelakantan, Pranav Shyam, Girish Sastry, Amanda
  Askell, Sandhini Agarwal, Ariel Herbert{-}Voss, Gretchen Krueger, Tom
  Henighan, Rewon Child, Aditya Ramesh, Daniel~M. Ziegler, Jeffrey Wu, Clemens
  Winter, Christopher Hesse, Mark Chen, Eric Sigler, Mateusz Litwin, Scott
  Gray, Benjamin Chess, Jack Clark, Christopher Berner, Sam McCandlish, Alec
  Radford, Ilya Sutskever, and Dario Amodei.
\newblock Language models are few-shot learners.
\newblock \emph{CoRR}, abs/2005.14165, 2020.
\newblock URL \url{https://arxiv.org/abs/2005.14165}.

\bibitem[Chollet(2017)]{chollet2017xception}
Fran{\c{c}}ois Chollet.
\newblock Xception: Deep learning with depthwise separable convolutions.
\newblock In \emph{Proceedings of the IEEE conference on computer vision and
  pattern recognition}, pp.\  1251--1258, 2017.

\bibitem[Ding et~al.(2022)Ding, Zheng, Hong, and Tang]{ding2022cogview2}
Ming Ding, Wendi Zheng, Wenyi Hong, and Jie Tang.
\newblock Cogview2: Faster and better text-to-image generation via hierarchical
  transformers.
\newblock \emph{arXiv preprint arXiv:2204.14217}, 2022.

\bibitem[Gafni et~al.(2022)Gafni, Polyak, Ashual, Sheynin, Parikh, and
  Taigman]{makeascene}
Oran Gafni, Adam Polyak, Oron Ashual, Shelly Sheynin, Devi Parikh, and Yaniv
  Taigman.
\newblock Make-a-scene: Scene-based text-to-image generation with human priors,
  2022.
\newblock URL \url{https://arxiv.org/abs/2203.13131}.

\bibitem[Ge et~al.(2022)Ge, Hayes, Yang, Yin, Pang, Jacobs, Huang, and
  Parikh]{ge2022long}
Songwei Ge, Thomas Hayes, Harry Yang, Xi~Yin, Guan Pang, David Jacobs, Jia-Bin
  Huang, and Devi Parikh.
\newblock Long video generation with time-agnostic vqgan and time-sensitive
  transformer.
\newblock \emph{ECCV}, 2022.

\bibitem[Girish et~al.(2020)Girish, Singh, and Ralescu]{actionsurvey2}
Deeptha Girish, Vineeta Singh, and Anca Ralescu.
\newblock Understanding action recognition in still images.
\newblock pp.\  1523--1529, 06 2020.
\newblock \doi{10.1109/CVPRW50498.2020.00193}.

\bibitem[Goodfellow et~al.(2014)Goodfellow, Pouget-Abadie, Mirza, Xu,
  Warde-Farley, Ozair, Courville, and Bengio]{goodfellow2014generative}
Ian Goodfellow, Jean Pouget-Abadie, Mehdi Mirza, Bing Xu, David Warde-Farley,
  Sherjil Ozair, Aaron Courville, and Yoshua Bengio.
\newblock Generative adversarial networks.
\newblock \emph{NIPS}, 2014.

\bibitem[Gu et~al.(2022)Gu, Chen, Bao, Wen, Zhang, Chen, Yuan, and
  Guo]{gu2022vector}
Shuyang Gu, Dong Chen, Jianmin Bao, Fang Wen, Bo~Zhang, Dongdong Chen, Lu~Yuan,
  and Baining Guo.
\newblock Vector quantized diffusion model for text-to-image synthesis.
\newblock In \emph{CVPR}, pp.\  10696--10706, 2022.

\bibitem[Gupta et~al.(2018)Gupta, Schwenk, Farhadi, Hoiem, and
  Kembhavi]{gupta2018imagine}
Tanmay Gupta, Dustin Schwenk, Ali Farhadi, Derek Hoiem, and Aniruddha Kembhavi.
\newblock Imagine this! scripts to compositions to videos.
\newblock In \emph{ECCV}, pp.\  598--613, 2018.

\bibitem[Hayes et~al.(2022)Hayes, Zhang, Yin, Pang, Sheng, Yang, Ge, Hu, and
  Parikh]{hayes2022mugen}
Thomas Hayes, Songyang Zhang, Xi~Yin, Guan Pang, Sasha Sheng, Harry Yang,
  Songwei Ge, Isabelle Hu, and Devi Parikh.
\newblock Mugen: A playground for video-audio-text multimodal understanding and
  generation.
\newblock \emph{ECCV}, 2022.

\bibitem[Ho et~al.(2020)Ho, Jain, and Abbeel]{ddpm}
Jonathan Ho, Ajay Jain, and Pieter Abbeel.
\newblock Denoising diffusion probabilistic models, 2020.
\newblock URL \url{https://arxiv.org/abs/2006.11239}.

\bibitem[Ho et~al.(2022)Ho, Salimans, Gritsenko, Chan, Norouzi, and Fleet]{VDM}
Jonathan Ho, Tim Salimans, Alexey Gritsenko, William Chan, Mohammad Norouzi,
  and David~J. Fleet.
\newblock Video diffusion models, 2022.
\newblock URL \url{https://arxiv.org/abs/2204.03458}.

\bibitem[Hong et~al.(2018)Hong, Yang, Choi, and Lee]{hong2018inferring}
Seunghoon Hong, Dingdong Yang, Jongwook Choi, and Honglak Lee.
\newblock Inferring semantic layout for hierarchical text-to-image synthesis.
\newblock In \emph{CVPR}, pp.\  7986--7994, 2018.

\bibitem[Hong et~al.(2022)Hong, Ding, Zheng, Liu, and Tang]{CogVideo}
Wenyi Hong, Ming Ding, Wendi Zheng, Xinghan Liu, and Jie Tang.
\newblock Cogvideo: Large-scale pretraining for text-to-video generation via
  transformers, 2022.
\newblock URL \url{https://arxiv.org/abs/2205.15868}.

\bibitem[Li et~al.(2018)Li, Min, Shen, Carlson, and Carin]{li2018video}
Yitong Li, Martin Min, Dinghan Shen, David Carlson, and Lawrence Carin.
\newblock Video generation from text.
\newblock In \emph{AAAI}, volume~32, 2018.

\bibitem[Liu et~al.(2019{\natexlab{a}})Liu, Ott, Goyal, Du, Joshi, Chen, Levy,
  Lewis, Zettlemoyer, and Stoyanov]{roberta}
Yinhan Liu, Myle Ott, Naman Goyal, Jingfei Du, Mandar Joshi, Danqi Chen, Omer
  Levy, Mike Lewis, Luke Zettlemoyer, and Veselin Stoyanov.
\newblock Roberta: {A} robustly optimized {BERT} pretraining approach.
\newblock \emph{CoRR}, abs/1907.11692, 2019{\natexlab{a}}.
\newblock URL \url{http://arxiv.org/abs/1907.11692}.

\bibitem[Liu et~al.(2019{\natexlab{b}})Liu, Wang, Yuan, and Zhu]{liu2019cross}
Yue Liu, Xin Wang, Yitian Yuan, and Wenwu Zhu.
\newblock Cross-modal dual learning for sentence-to-video generation.
\newblock In \emph{Proceedings of the 27th ACM International Conference on
  Multimedia}, pp.\  1239--1247, 2019{\natexlab{b}}.

\bibitem[Marwah et~al.(2017)Marwah, Mittal, and
  Balasubramanian]{marwah2017attentive}
Tanya Marwah, Gaurav Mittal, and Vineeth~N Balasubramanian.
\newblock Attentive semantic video generation using captions.
\newblock In \emph{ICCV}, pp.\  1426--1434, 2017.

\bibitem[Mittal et~al.(2017)Mittal, Marwah, and
  Balasubramanian]{mittal2017sync}
Gaurav Mittal, Tanya Marwah, and Vineeth~N Balasubramanian.
\newblock Sync-draw: Automatic video generation using deep recurrent attentive
  architectures.
\newblock In \emph{Proceedings of the 25th ACM international conference on
  Multimedia}, pp.\  1096--1104, 2017.

\bibitem[Nichol et~al.(2021)Nichol, Dhariwal, Ramesh, Shyam, Mishkin, McGrew,
  Sutskever, and Chen]{nichol2021glide}
Alex Nichol, Prafulla Dhariwal, Aditya Ramesh, Pranav Shyam, Pamela Mishkin,
  Bob McGrew, Ilya Sutskever, and Mark Chen.
\newblock Glide: Towards photorealistic image generation and editing with
  text-guided diffusion models.
\newblock \emph{arXiv preprint arXiv:2112.10741}, 2021.

\bibitem[Pan et~al.(2017)Pan, Qiu, Yao, Li, and Mei]{pan2017create}
Yingwei Pan, Zhaofan Qiu, Ting Yao, Houqiang Li, and Tao Mei.
\newblock To create what you tell: Generating videos from captions.
\newblock In \emph{Proceedings of the 25th ACM international conference on
  Multimedia}, pp.\  1789--1798, 2017.

\bibitem[Parmar et~al.(2022)Parmar, Zhang, and Zhu]{parmar2021cleanfid}
Gaurav Parmar, Richard Zhang, and Jun-Yan Zhu.
\newblock On aliased resizing and surprising subtleties in gan evaluation.
\newblock In \emph{CVPR}, 2022.

\bibitem[Qiu et~al.(2017)Qiu, Yao, and Mei]{qiu2017learning}
Zhaofan Qiu, Ting Yao, and Tao Mei.
\newblock Learning spatio-temporal representation with pseudo-3d residual
  networks.
\newblock In \emph{ICCV}, pp.\  5533--5541, 2017.

\bibitem[Radford et~al.(2021)Radford, Kim, Hallacy, Ramesh, Goh, Agarwal,
  Sastry, Askell, Mishkin, Clark, et~al.]{radford2021learning}
Alec Radford, Jong~Wook Kim, Chris Hallacy, Aditya Ramesh, Gabriel Goh,
  Sandhini Agarwal, Girish Sastry, Amanda Askell, Pamela Mishkin, Jack Clark,
  et~al.
\newblock Learning transferable visual models from natural language
  supervision.
\newblock In \emph{ICML}, pp.\  8748--8763. PMLR, 2021.

\bibitem[Ramesh et~al.(2021)Ramesh, Pavlov, Goh, Gray, Voss, Radford, Chen, and
  Sutskever]{ramesh2021zero}
Aditya Ramesh, Mikhail Pavlov, Gabriel Goh, Scott Gray, Chelsea Voss, Alec
  Radford, Mark Chen, and Ilya Sutskever.
\newblock Zero-shot text-to-image generation.
\newblock In \emph{ICML}, pp.\  8821--8831. PMLR, 2021.

\bibitem[Ramesh et~al.(2022)Ramesh, Dhariwal, Nichol, Chu, and Chen]{dalle2}
Aditya Ramesh, Prafulla Dhariwal, Alex Nichol, Casey Chu, and Mark Chen.
\newblock Hierarchical text-conditional image generation with clip latents,
  2022.
\newblock URL \url{https://arxiv.org/abs/2204.06125}.

\bibitem[Reda et~al.(2022)Reda, Kontkanen, Tabellion, Sun, Pantofaru, and
  Curless]{reda2022film}
Fitsum Reda, Janne Kontkanen, Eric Tabellion, Deqing Sun, Caroline Pantofaru,
  and Brian Curless.
\newblock Film: Frame interpolation for large motion.
\newblock \emph{arXiv preprint arXiv:2202.04901}, 2022.

\bibitem[Reed et~al.(2016)Reed, Akata, Yan, Logeswaran, Schiele, and
  Lee]{reed2016generative}
Scott Reed, Zeynep Akata, Xinchen Yan, Lajanugen Logeswaran, Bernt Schiele, and
  Honglak Lee.
\newblock Generative adversarial text to image synthesis.
\newblock In \emph{ICML}, pp.\  1060--1069. PMLR, 2016.

\bibitem[Rombach et~al.(2022)Rombach, Blattmann, Lorenz, Esser, and
  Ommer]{rombach2022high}
Robin Rombach, Andreas Blattmann, Dominik Lorenz, Patrick Esser, and Bj{\"o}rn
  Ommer.
\newblock High-resolution image synthesis with latent diffusion models.
\newblock In \emph{CVPR}, pp.\  10684--10695, 2022.

\bibitem[Saharia et~al.(2022)Saharia, Chan, Saxena, Li, Whang, Denton,
  Ghasemipour, Ayan, Mahdavi, Lopes, Salimans, Ho, Fleet, and Norouzi]{imagen}
Chitwan Saharia, William Chan, Saurabh Saxena, Lala Li, Jay Whang, Emily
  Denton, Seyed Kamyar~Seyed Ghasemipour, Burcu~Karagol Ayan, S.~Sara Mahdavi,
  Rapha~Gontijo Lopes, Tim Salimans, Jonathan Ho, David~J Fleet, and Mohammad
  Norouzi.
\newblock Photorealistic text-to-image diffusion models with deep language
  understanding, 2022.
\newblock URL \url{https://arxiv.org/abs/2205.11487}.

\bibitem[Saito et~al.(2020)Saito, Saito, Koyama, and Kobayashi]{saito2020train}
Masaki Saito, Shunta Saito, Masanori Koyama, and Sosuke Kobayashi.
\newblock Train sparsely, generate densely: Memory-efficient unsupervised
  training of high-resolution temporal gan.
\newblock \emph{International Journal of Computer Vision}, 128\penalty0
  (10):\penalty0 2586--2606, 2020.

\bibitem[Schuhmann et~al.()Schuhmann, Beaumont, Gordon, Wightman, Coombes,
  Katta, Mullis, Schramowski, Kundurthy, Crowson, et~al.]{schuhmannlaion}
Christoph Schuhmann, Romain Beaumont, Cade~W Gordon, Ross Wightman, Theo
  Coombes, Aarush Katta, Clayton Mullis, Patrick Schramowski, Srivatsa~R
  Kundurthy, Katherine Crowson, et~al.
\newblock Laion-5b: An open large-scale dataset for training next generation
  image-text models.

\bibitem[Schuhmann et~al.(2022)Schuhmann, Vencu, Beaumont, Coombes, Gordon,
  Katta, Kaczmarczyk, and Jitsev]{laion5b}
Christoph Schuhmann, Richard Vencu, Romain Beaumont, Theo Coombes, Cade Gordon,
  Aarush Katta, Robert Kaczmarczyk, and Jenia Jitsev.
\newblock {LAION-5B:} laion-5b: A new era of open large-scale multi-modal
  datasets.
\newblock
  \url{https://laion.ai/laion-5b-a-new-era-of-open-large-scale-multi-modal-datasets/},
  2022.

\bibitem[Soomro et~al.(2012)Soomro, Zamir, and Shah]{soomro2012ucf101}
Khurram Soomro, Amir~Roshan Zamir, and Mubarak Shah.
\newblock Ucf101: A dataset of 101 human actions classes from videos in the
  wild.
\newblock \emph{arXiv preprint arXiv:1212.0402}, 2012.

\bibitem[Tian et~al.(2021)Tian, Ren, Chai, Olszewski, Peng, Metaxas, and
  Tulyakov]{tian2021good}
Yu~Tian, Jian Ren, Menglei Chai, Kyle Olszewski, Xi~Peng, Dimitris~N Metaxas,
  and Sergey Tulyakov.
\newblock A good image generator is what you need for high-resolution video
  synthesis.
\newblock \emph{ICLR}, 2021.

\bibitem[Vaswani et~al.(2017)Vaswani, Shazeer, Parmar, Uszkoreit, Jones, Gomez,
  Kaiser, and Polosukhin]{transformer}
Ashish Vaswani, Noam Shazeer, Niki Parmar, Jakob Uszkoreit, Llion Jones,
  Aidan~N. Gomez, Lukasz Kaiser, and Illia Polosukhin.
\newblock Attention is all you need, 2017.
\newblock URL \url{https://arxiv.org/abs/1706.03762}.

\bibitem[Wu et~al.(2021{\natexlab{a}})Wu, Huang, Zhang, Li, Ji, Yang, Sapiro,
  and Duan]{wu2021godiva}
Chenfei Wu, Lun Huang, Qianxi Zhang, Binyang Li, Lei Ji, Fan Yang, Guillermo
  Sapiro, and Nan Duan.
\newblock Godiva: Generating open-domain videos from natural descriptions.
\newblock \emph{arXiv preprint arXiv:2104.14806}, 2021{\natexlab{a}}.

\bibitem[Wu et~al.(2021{\natexlab{b}})Wu, Liang, Ji, Yang, Fang, Jiang, and
  Duan]{nuwa}
Chenfei Wu, Jian Liang, Lei Ji, Fan Yang, Yuejian Fang, Daxin Jiang, and Nan
  Duan.
\newblock NÜwa: Visual synthesis pre-training for neural visual world
  creation, 2021{\natexlab{b}}.
\newblock URL \url{https://arxiv.org/abs/2111.12417}.

\bibitem[Xie et~al.(2018)Xie, Sun, Huang, Tu, and Murphy]{xie2018rethinking}
Saining Xie, Chen Sun, Jonathan Huang, Zhuowen Tu, and Kevin Murphy.
\newblock Rethinking spatiotemporal feature learning: Speed-accuracy trade-offs
  in video classification.
\newblock In \emph{ECCV}, pp.\  305--321, 2018.

\bibitem[Xu et~al.(2016)Xu, Mei, Yao, and Rui]{xu2016msr}
Jun Xu, Tao Mei, Ting Yao, and Yong Rui.
\newblock Msr-vtt: A large video description dataset for bridging video and
  language.
\newblock In \emph{CVPR}, pp.\  5288--5296, 2016.

\bibitem[Xu et~al.(2018)Xu, Zhang, Huang, Zhang, Gan, Huang, and
  He]{xu2018attngan}
Tao Xu, Pengchuan Zhang, Qiuyuan Huang, Han Zhang, Zhe Gan, Xiaolei Huang, and
  Xiaodong He.
\newblock Attngan: Fine-grained text to image generation with attentional
  generative adversarial networks.
\newblock In \emph{CVPR}, pp.\  1316--1324, 2018.

\bibitem[Xue et~al.(2022)Xue, Hang, Zeng, Sun, Liu, Yang, Fu, and
  Guo]{xue2022advancing}
Hongwei Xue, Tiankai Hang, Yanhong Zeng, Yuchong Sun, Bei Liu, Huan Yang,
  Jianlong Fu, and Baining Guo.
\newblock Advancing high-resolution video-language representation with
  large-scale video transcriptions.
\newblock In \emph{CVPR}, pp.\  5036--5045, 2022.

\bibitem[Ye et~al.(2019)Ye, Liu, and Zhang]{ye20193d}
Rongtian Ye, Fangyu Liu, and Liqiang Zhang.
\newblock 3d depthwise convolution: Reducing model parameters in 3d vision
  tasks.
\newblock In \emph{Canadian Conference on Artificial Intelligence}, pp.\
  186--199. Springer, 2019.

\bibitem[Yu et~al.(2021)Yu, Li, Koh, Zhang, Pang, Qin, Ku, Xu, Baldridge, and
  Wu]{yu2021vector}
Jiahui Yu, Xin Li, Jing~Yu Koh, Han Zhang, Ruoming Pang, James Qin, Alexander
  Ku, Yuanzhong Xu, Jason Baldridge, and Yonghui Wu.
\newblock Vector-quantized image modeling with improved vqgan.
\newblock \emph{arXiv preprint arXiv:2110.04627}, 2021.

\bibitem[Yu et~al.(2022{\natexlab{a}})Yu, Xu, Koh, Luong, Baid, Wang,
  Vasudevan, Ku, Yang, Ayan, Hutchinson, Han, Parekh, Li, Zhang, Baldridge, and
  Wu]{parti}
Jiahui Yu, Yuanzhong Xu, Jing~Yu Koh, Thang Luong, Gunjan Baid, Zirui Wang,
  Vijay Vasudevan, Alexander Ku, Yinfei Yang, Burcu~Karagol Ayan, Ben
  Hutchinson, Wei Han, Zarana Parekh, Xin Li, Han Zhang, Jason Baldridge, and
  Yonghui Wu.
\newblock Scaling autoregressive models for content-rich text-to-image
  generation, 2022{\natexlab{a}}.
\newblock URL \url{https://arxiv.org/abs/2206.10789}.

\bibitem[Yu et~al.(2022{\natexlab{b}})Yu, Tack, Mo, Kim, Kim, Ha, and
  Shin]{yu2022generating}
Sihyun Yu, Jihoon Tack, Sangwoo Mo, Hyunsu Kim, Junho Kim, Jung-Woo Ha, and
  Jinwoo Shin.
\newblock Generating videos with dynamics-aware implicit generative adversarial
  networks.
\newblock \emph{ICLR}, 2022{\natexlab{b}}.

\bibitem[Zhang et~al.(2017)Zhang, Xu, Li, Zhang, Wang, Huang, and
  Metaxas]{zhang2017stackgan}
Han Zhang, Tao Xu, Hongsheng Li, Shaoting Zhang, Xiaogang Wang, Xiaolei Huang,
  and Dimitris~N Metaxas.
\newblock Stackgan: Text to photo-realistic image synthesis with stacked
  generative adversarial networks.
\newblock In \emph{ICCV}, pp.\  5907--5915, 2017.

\bibitem[Zhang et~al.(2021)Zhang, Koh, Baldridge, Lee, and
  Yang]{zhang2021cross}
Han Zhang, Jing~Yu Koh, Jason Baldridge, Honglak Lee, and Yinfei Yang.
\newblock Cross-modal contrastive learning for text-to-image generation.
\newblock In \emph{CVPR}, pp.\  833--842, 2021.

\end{thebibliography}
\bibliographystyle{paper}

\end{document}